%% file: main.tex
\documentclass[10pt,twocolumn,letterpaper]{article}

\usepackage{cvpr}
\usepackage{times}
\usepackage{epsfig}
\usepackage{graphicx}
\usepackage{amsmath}
\usepackage{amssymb}
\usepackage{placeins}
\usepackage{color}

\newcommand{\fig}[1]{Figure~\ref{fig:#1}}
\newcommand{\sect}[1]{Section~\ref{sect:#1}}

\newcommand{\eq}[1]{(\ref{eq:#1})}

\newcommand{\p}{\mathcal{P}}
\newcommand{\e}{e}
\newcommand{\E}{E}

\usepackage[pagebackref=true,breaklinks=true,letterpaper=true,colorlinks,bookmarks=false]{hyperref}

\cvprfinalcopy % *** Uncomment this line for the final submission

\def\cvprPaperID{2782} % *** Enter the CVPR Paper ID here

\ifcvprfinal\fi

\begin{document}

%%%%%%%%% TITLE
\title{Parsing Images of Overlapping Organisms with\\Deep Singling-Out Networks} 

\author{Victor Yurchenko\\
Skoltech\\
Russia\\
{\tt\small victor.yurchenko@skoltech.ru}
\and
Victor Lempitsky\\
Skoltech\\
Russia\\
{\tt\small lempitsky@skoltech.ru}
}

\maketitle
\begin{abstract}
This work is motivated by the mostly unsolved task of parsing biological images with multiple overlapping articulated model organisms (such as worms or larvae). We present a general approach that separates the two main challenges associated with such data, individual object shape estimation and object groups disentangling. At the core of the approach is a deep feed-forward \textit{singling-out network (SON)} that is trained to map each local patch to a vectorial descriptor that is sensitive to the characteristics (e.g.\ shape) of a central object, while being invariant to the variability of all other surrounding elements. Given a SON, a local image patch can be matched to a gallery of isolated elements using their SON-descriptors, thus producing a hypothesis about the shape of the central element in that patch. The image-level optimization based on integer programming can then pick a subset of the hypotheses to explain (parse) the whole image and disentangle groups of organisms. 

While sharing many similarities with existing ``analysis-by-synthesis'' approaches, our method avoids the need for stochastic search in the high-dimensional configuration space and numerous rendering operations at test-time. We show that our approach can parse microscopy images of three popular model organisms (the C.Elegans roundworms, the Drosophila larvae, and the E.\ Coli bacteria) even under significant crowding and overlaps between organisms. We speculate that the overall approach is applicable to a wider class of image parsing problems concerned with crowded articulated objects, for which rendering training images is possible. 

%\keywords{Biomedical image analysis, deep learning, image parsing, instance segmentation, integer programming}
\end{abstract}

\input{intro}

\input{related}

\input{method}

\input{experiments}
\input{conclusion}

%\section*{Acknowledgements}

\FloatBarrier
{\small
\bibliographystyle{ieee}
\bibliography{refs}
}
\end{document}

%% file: intro.tex
\section{Introduction}

\setlength{\tabcolsep}{3pt}

\begin{figure}
    \centering
    \begin{tabular}{ccc}
    \includegraphics[height=2.5cm]{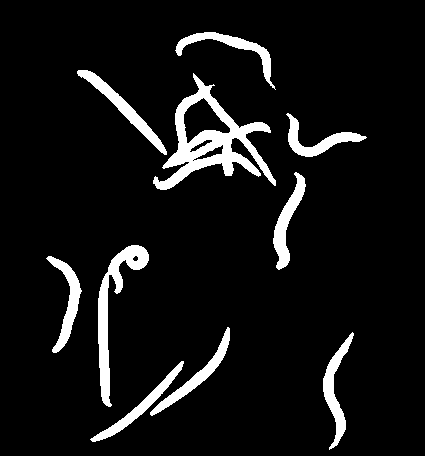}&
    \includegraphics[height=2.5cm]{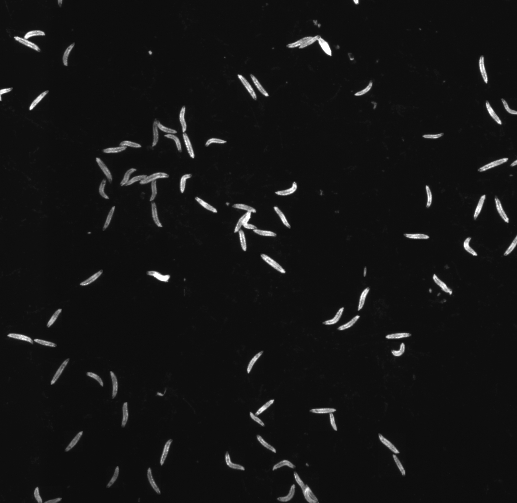}&
    \includegraphics[height=2.5cm]{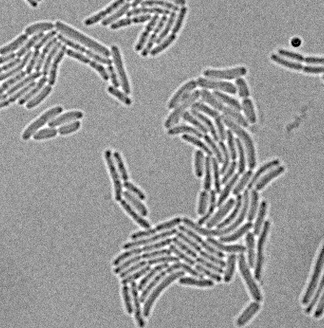}\\
    \vspace{-1mm} \textit{C Elegans}& \textit{Drosophila} &  \textit{E.Coli}\\
     (roundworms) & (larvae) &  (bacteria)
    \end{tabular}
    
    \caption{We consider the image parsing tasks for three different organisms that are popular in biomedical research. In each case, parsing is made hard because of a certain shape variability of individual organisms as well as organism overlap and crowding. Although the three organisms are very different biologically, we approach the corresponding parsing tasks with a unified framework that first uses a specially-designed deep network to propose hypotheses about the shapes of individual organisms and then use integer programming to pick a viable hypotheses set.}
    \label{fig:input}
\end{figure}

Parsing images of biological substances has become one of the important applications of computer vision \cite{Eliceiri12}. In many biologically-important scenarios it is necessary to deal with images of overlapping objects or organisms. In recent years, several approaches have been proposed that can parse images when objects have simple blob-type shapes (e.g.\ cell cultures) \cite{AlKofani10,Arteta16,Bernardis10,Descombes09,Zhang14}. Less attention, however, has been paid to images containing more complex organisms exhibiting significant shape and pose variations, such as worms, larvae, and bacilli. The sheer importance of such model organisms for biomedical studies calls for further improvement of parsing approaches for this class of images. 

Two factors make parsing of such images a complicated task. First, organisms can exhibit significant rigid and non-rigid pose variations. Secondly, these organisms often form \textit{clusters} that cannot be segmented into individual organisms using simple image processing methods. The two factors complicate each other, as the variation of the appearance of clusters can be combinatorially larger than the variation of the appearance of a single organism, thus defying brute-force parsing approaches.

\begin{figure}
\centering

\newlength{\lenm}
\setlength{\lenm}{1.3cm}

\newcommand{\incmatch}[1]{
\includegraphics[width=\lenm]{figures/match/#1_input.png}&
\includegraphics[width=\lenm]{figures/match/#1_nn0.png}&
\includegraphics[width=\lenm]{figures/match/#1_nn1.png}&
\includegraphics[width=\lenm]{figures/match/#1_nn2.png}\\
}
\begin{tabular}{cccc}
\incmatch{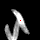}
%\incmatch{123}
\incmatch{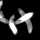}
\incmatch{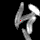}
%\incmatch{399}
%\incmatch{490}
%\incmatch{495}
%\incmatch{887}
\end{tabular}
\caption{Given a patch containing overlapping organisms (here Drosophila larvae), our deep architecture (the \textit{SON-network}) computes a vectorial \textit{SON-descriptor}. We then perform nearest-neighbor search in the gallery of images of single organisms with precomputed SON-descriptors (here, images corresponding to the first three nearest neighbors are shown). Because of the properties of these descriptors, the matched organisms have similar shapes/poses to the organism that covers the central pixel of the query patch. The remaining organisms in the query patch have little effect on the matching process. The recovered hypotheses about central organisms can be then used in the whole image parsing process.}
\label{fig:teaser}
\end{figure}

Our approach uses a combination of deep learning and generative modeling to tackle the challenge of organism cluster parsing. The approach starts by training a deep feed-forward network that maps each local image patch $\p$ to a descriptor that is sensitive to the configuration of the central object in $\p$, while being insensitive to other objects in  $\p$. Informally speaking, such a \textit{singling-out network} (SON), distinguishes the central element from its surrounding, and then describes the appearance/configuration of this element by a  \textit{SON-descriptor}. At test-time, the SON-network allows to obtain a large set of hypotheses about individual objects in the cluster. This is done by comparing SON-descriptors of various image patches covering the cluster against a pre-computed large set of SON-descriptors of patches with known central elements (\fig{teaser}). As a last step, we use a facility-location type discrete optimization~\cite{Lazic09,Barinova12,Delong12} to pick a small subset of hypotheses that ``explains'' the appearance of the whole cluster.

Below, in \sect{experiments} we show that this approach can be successfully applied to three diverse datasets corresponding to three popular model organisms: C.Elegans roundworm, Drosophila larva, and E.Coli bacterium (\fig{input}). Before that, we discuss prior related work in \sect{related}, and then explain our method in \sect{method}. We conclude by a short discussion in \sect{discussion}.

%% file: related.tex
\section{Related work}
\label{sect:related}

Despite large practical importance, there is little published work dedicated to the image analysis task we focus on. W{\"{a}}hlby~et~al.~\cite{Wahlby10} describe a method for resolving C.Elegans worms clusters based on probabilistic analysis, which achieves impressive results. It however makes several assumptions specific to particular organism/assay types that can be potentially brittle, such as the ability to isolate tips of organisms or the ability to mine worm centerlines as paths in the cluster skeleton. More recently, Fiaschi~et~al.~\cite{Fiaschi13,Fiaschi14} addressed the problem of Drosophila larvae tracking through network and integer programming. Our approach is quite different to theirs, as we focus on handling single frames. Below, we present results for both W{\"{a}}hlby~et~al. and Fiaschi~et~al. data obtained with our method.

Algorithmically, our approach builds upon two streams of ideas. The first stream are methods based on deep discriminatively-trained deep convolutional networks~\cite{LeCun89}, which currently enjoy overwhelming success in image analysis. Here, the components of our methods resembles the combination of deep descriptors and nearest-neighbor search in \cite{Ganin14}.  

The second relevant group of methods is formed by generative ``analysis-by-synthesis'' frameworks \cite{Descombes09,Kulkarni15,Verdie12}. A recent work of Kulkarni et al.~\cite{Kulkarni15} nicely combines the two streams by using deep features to compare the synthesized and the input images. ``Analysis-by-synthesis'' approaches are appealing due to their conceptual simplicity, and overall hold great potential. However the complexity of scenes that they can parse is limited by the need to perform stochastic search over the scene configuration space and the need to re-render the scene at each step of such search. These computational hurdles are avoided in our method.

Analyzing crowded scenes by suggesting an excessive number of hypotheses and then picking a subset of them through optimization is an idea that has been used in several computer vision works. For example, Wu~and~Nevatia~\cite{Wu07} used edge-based human part detectors to hypothesize about individual locations in crowded surveillance videos. Likewise, \cite{Barinova12} used discriminatively-trained Hough transform to obtain hypotheses. In both cases, greedy optimization was used to pick optimal subsets, but other optimization methods could have been used. Compared to this group of methods, our contribution is the specific way the hypotheses are obtained (SON-networks).

% , later picking subset through analysis-by-synthesis to parse images of overlapping humans using responces of edge-based parts detectors to compare suggested model and input images. In \cite{Barinova10}, crowded scenes are analyzed by identifying hypotheses through discriminatively-trained generalized Hough transform, and then picking a subset of those via facility-location type optimization. This  

% Walhby et al.

% Methods that parse blob cells

% Deep descriptors, Chopra, faces, N4-fields

% Analysis by synthesis Kulkarni, Nevatia

% Detectors of specific occlusion patterns Tang et al. bmvc12, 

% Facility location in computer vision

%% file: method.tex
\section{Method}
\label{sect:method}

In a nutshell, our approach focuses on (partial) understanding of image patches, and then integrating the information from individual patches into a holistic image interpretation via a joint optimization process.

Let us first introduce the notation at the level of a certain patch $\p$. In most microscopy image parsing scenarios (including ours) the binary object/background segmentation is relatively easy, and therefore it is easy to discard patches where the central pixels are not covered by the foreground elements. We therefore restrict our attention to the remaining patches. We thus assume that the patch has a set of objects (elements) $\E^\p = \{\e^\p_0,\e^\p_1,\dots \e^\p_{N_\p}\}$ overlapping with it, and that $\e^\p_0$ denotes the  \textit{central element} that covers the center of the patch. Generally, we assume that each element $\e^\p_i$ is characterized by several degrees of freedom (e.g.\ center position, orientation, shape parameters, texture parameters).

We denote with $I(\p)$ the appearance of the patch (a multichannel image of a certain size), and assume that $I(\p) = R(\E^\p; \xi)$, where $R$ is the \textit{rendering function}, and $\xi$ is a nuisance variable that incorporates such factors as image noise or some clutter that we are not aiming to recover, etc. We further assume that we have a reasonable approximation of the rendering function $R$, and that we can draw samples from the distribution of elements.

\subsection{Inverse rendering using Singling-Out Networks}

The key idea of our approach is to learn the \textit{partial inverse} mapping $S: I(\p) \to \e^\p_0$ that recovers the central element  $\e^\p_0$ from the appearance $I(\p)$ while ignoring the impact of $\e^\p_1,\dots \e^\p_{N_\p}$ on the appearance $I$. Overall, we achieve this using the combination of a deep feedforward network learning and nearest-neighbor search.

The vital component of such partial reverse mapping is a deep feedforward \textit{singling-out} network $f(I;\Theta)$ that maps the appearance $I$ of an image patch to a high-dimensional descriptor vector $d$ (where $\Theta$ are the parameters of the deep network). The learning process tries to adjust $\Theta$ to ensure that the appearance of patches with similar central elements are mapped to close \textit{singling-out network (SON)} descriptor vectors and vice versa. 

\newlength{\lensev}
\setlength{\lensev}{1.25cm}

\newcommand{\incclass}[1]{
\includegraphics[width=\lensev]{figures/classes/#1/0.png}&
\includegraphics[width=\lensev]{figures/classes/#1/1.png}&
\includegraphics[width=\lensev]{figures/classes/#1/2.png}&
\includegraphics[width=\lensev]{figures/classes/#1/3.png}&
\includegraphics[width=\lensev]{figures/classes/#1/6.png}\\
}

\begin{figure}
    \centering
    \begin{tabular}{ccccc}
    \incclass{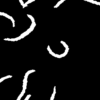}
    \incclass{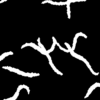}
    \end{tabular}
    
    \caption{The classes needed to train SON-network are generated as sets of images with the same organism (here C.Elegans) in the center. Here, each row shows several synthetic examples from the same class. The network then has to learn features that can help it to distinguish (``single out'') the central organism from the occluders.}
    \label{fig:classes}
\end{figure}

{\bf Training SON-networks.} There are several potential approaches to the training process of the SON-networks. One can use Siamese pairwise loss \cite{Chopra05} or triplet loss \cite{Schroff15}, which would require sampling pairs or triplets of patches with some patches having ``similar'' central elements, and other patches having ``dissimilar'' central elements. Pair-based and triplet-based learning of deep feedforward networks is however known to be hard in terms of finding suitable initialization, setting the meta-parameters of the network architecture and the learning process (learning rate), as well as setting pairs/triplets generation properly. Therefore we used a proxy classification problem (\fig{classes}) to learn a classification network (as in e.g.\ \cite{Taigman14}) using a standard classification softmax loss.

In the training classification dataset, each class $j$ is generated as follows. At first, a random central element $\e^j_0$ is drawn. Then each training image of the class is created by sampling additional elements $\e^j_{i,1}, \e^j_{i,2}, \dots, \e^j_{i,n_{j,i}}$ and a random nuisance parameter $\xi^j_i$ and rendering the correspondence appearance:
\begin{equation}
I^j_i = R\left(\{e^j_0,\e^j_{i,1}, \e^j_{i,2}, \dots, \e^j_{i,n_{j,i}}\}, \xi^j_i\right)
\end{equation}
The training class $j$ then consists of images $I^j_i$ for all possible $i$.

The SON-network is then trained to classify between a large number of classes generated with this procedure. In our experiments, we use convolutional neural networks \cite{LeCun89} with three convolutional and three fully-connected layers. After training the last layer that predicts class posteriors is discarded and the output of the penultimate layer serves as a descriptor of the input image (i.e.\ the feedforward mapping from the input image to the activations of the penultimate layer serve as $f(I;\Theta)$).

{\bf Gallery matching.} We augment the deep descriptor learning with nearest-neighbor search to conclude the partial inverse mapping. We thus synthesize $K$ random central elements $\hat{e}_1,\hat{e}_2,\dots \hat{e}_K$, render them, and then pass the resulting image patches through the trained SON-network, obtaining their SON-descriptors $\hat{d}_i$:
\begin{equation}
\hat{d}_i = f\left(R(\{\hat{e}_i\};\xi_i);\Theta\right)
\end{equation}
The elements together with their descriptors are then stored in a \textit{gallery} $\{\hat{d}_1{:}\hat{e}_1,\hat{d}_2{:}\hat{e}_2,\dots,\hat{d}_K{:}\hat{e}_K\}$. Alternatively to artificial rendering process, the gallery patches can be sampled from the annotated training images, whereas geometric and photometric data augmentation can be used to increase the diversity of gallery patches.

Given an image patch $I$ we can then generate a hypothesis about the central element in that patch by first obtaining its SON-descriptor $d=f(I;\Theta)$, then finding the nearest neighbor $\hat{d}_t$ in the gallery of SON-descriptors. The associated central element $\hat{e}_t$ then provides a hypothesis. Let us denote the compound mapping from the appearance $I$ to the hypothesis as $g$: $g(I) = \hat{e}_t$. \fig{teaser} provides the examples of such mapping.

\subsection{Image-level Parsing}

While the learned partial inverse mapping can provide a hypothesis for a single central patch, an additional optimization step is needed to obtain a set of elements that ``explain'' the entire image. 

To obtain the full image parsing, we first collect a set of hypotheses. For that, assuming that an approximate foreground/background segmentation is given, we consider a large number $M$ of patches with centers belonging to foreground. For each such patch centered at $(x_i,y_i)$ with the appearance $I_i$, we obtain a hypothesis $h_i$ using the partial inverse mapping, i.e.\ $h_i = g(I_i)$ (each hypothesis is thus just an element in a certain configuration). Since an element can be central for a number of patches (to be precise, for all patches centered at the pixels covered by the element), the set of hypotheses obtained in this way is excessive, and the goal of the further processing is to pick a subset of those.

We approach this pruning task using facility location-like optimization, which is a standard approach in image understanding (see e.g.\ \cite{Lazic09,Barinova12,Delong12}). We thus introduce binary variables $x_1,x_2,\dots x_M$, where $x_i=1$ means that the $i$-th hypothesis is selected ($x_i$ are thus ``facility'' variables). We demand that each of the $M$ patches we consider, is ``explained'' by one of the picked hypotheses (i.e.\ the patches are the ``clients''). To measure the quality of the explanation, we compute the value $d_{ij}$ that measures if the hypothesis $h_i$ can explain the patch $j$ (small values of $d_{ij}$ correspond to the case when such explanation is good).

The distance between the SON-descriptor of the patch $I_i$ and the SON-descriptor of the hypothesis $h_i$ is a natural choice for $d_{ii}$ as it is computed at the stage of the nearest-neighbor search:
\begin{equation} \label{eq:dii}
d_{ii} = \left\| f\left(I_i;\Theta\right) - f\left( R\left( (h_i); \xi\right) \right)  \right\|\;,
\end{equation}
where the nuisance parameter $\xi$ is taken arbitrarily.

One way to compute $d_{ij}$, when $i \ne j$ is to  evaluate the distance between the SON-descriptor of $I_j$ and the SON-descriptor of the hypothesis $h_i$ shifted according to the displacement between the $i$th and the $j$th pixel (in other words, in the coordinate frame associated with the patch $j$):
\begin{equation} \label{eq:dij1}
d_{ij} = \left\| f\left(I_j;\Theta\right) - f\left( R\left( T_{i\to j} (h_i); \xi\right) \right)  \right\|\;,
\end{equation}
where $T_{i\to j}$ is an operator that translates the element $h_i$ by $(x_j-x_i,y_j-y_i)$ into the coordinate frame associated with the $j$th patch before rendering.

Evaluating \eq{dij1} however requires rendering each hypothesis multiple times for different translations, and is therefore rather slow. Alternatively, for each gallery element one can precompute the change of the descriptor under different translations, and store it in the dataset. An easier approach is however to reuse the distance $d_{ii}$ computed in \eq{dii} for all patches with central pixels that are covered by the hypothesis $h_i$ placed on the image. Hence one can define:
\begin{equation} \label{eq:dij2}
d_{ij} = \delta(h_i,j)\,d_{ii},
\end{equation}
where $\delta(h_i,j)=1$ if the hypothesis $h_i$ covers the center of patch $p_j$ and $\delta(h_i,j)=+\infty$ otherwise (expressing the fact that the hypothesis cannot explain a patch, for which it does not cover the central pixel). Below, we present results for this fast approach using the distance estimates \eq{dij2}, and also selected results for the slow approach based on a more principled estimates \eq{dij1}.

Yet another fast heuristic that can be used to compute $d_{ij}$ is to look at the difference between the hypotheses suggested for the $i$th and the $j$th patches. In case, the two are covered by the same organism and the descriptor matching has worked well, the two hypotheses should be similar. Therefore, we can use some distance between hypotheses (e.g.\ the Hausdorf distance between hypotheses centerlines) to compute the distance estimates $d_{ij}$ (again the two hypotheses are compared in the ``global'' coordinate frames).

Once the distance estimates are computed, the binary variables $y_{ij}$ are introduced, where $y_{ij}=1$ means that the patch $j$ is actually explained by the hypothesis $h_i$ according to our image interpretation. The following optimization formulation (facility location) then implements the image parsing problem:
\begin{equation}
\begin{aligned}
& \underset{x,y}{\text{minimize}}
& & \sum_{i=1}^M \lambda\, x_i + \sum_{i=1}^M\sum_{j=1}^M d_{ij}\,y_{ij}\\
& \text{subject to}
& & x_i \in \{0,1\},\;\; y_{ij} \in \{0,1\} \\
&&& \forall i,j \;\; y_{ij} \le x_i, \\
&&& \forall i \;\; \sum_{j=1}^M y_{ij} =1.
\end{aligned} \label{eq:fl}
\end{equation}
Here, the first term in the objective implements the MDL (``minimum description length'') prior penalizing the number of selected hypotheses, whereas the coefficient $\lambda$ controls the strength of the prior, while the second term encourages the explanation of the patches by the hypotheses with matching SON-descriptors. 

The optimization of \eq{fl} is well studied in the context of computer vision application with a variety of problem-specific algorithms suggested \cite{Delong12}. We, however, utilize a general-purpose ILP solver~\cite{gurobi}, which allows solving \eq{fl} to global optimality in most cases in our experiments.

If it is known (e.g.\ from the training data) that two organisms cannot overlap beyond some threshold, we can find the set $S$ of all pairs of hypotheses $(k,l)$ such that $h_l$ and $h_k$ overlap by more than this threshold. We can then add the following constraint set into the optimization formulation \eq{fl}:
\begin{equation} \label{eq:overlap}
    \forall (k,l) \in S \quad x_k+x_l \le 1.
\end{equation}
Such set of equations ensures that too tightly overlapping hypotheses will not be picked simultaneously. While the set $S$ of such ``conflicting'' pairs can be very large, the constraints \eq{overlap} can be enforced in a cutting-plane fashion, starting the optimization without them and iteratively activating only the violated once while resolving for the optimal set. Due to the tendency of the facility location to avoid picking hypotheses that are too similar, few cutting plane iterations suffice in practice.

%% file: experiments.tex
\section{Experiments}
\label{sect:experiments}

\subsection{Datasets}
The \textbf{C.Elegans} dataset from the Broad Biomedical Benchmark Collection \cite{Ljosa12} consists of 100 images obtained using bright-field microscopy. The roundworms are subjected to various compounds, and the ultimate goal of the image processing in this case it to tell alive worms from the dead ones as dead worms possess characteristic (straight) shape. Similarly to \cite{Wahlby10}, we consider the binary segmentation rather than the raw data. We use 50 images for training, which leaves 50 images for testing. All images were scaled down by a factor of 2.5.

The \textbf{Larvae} dataset was used in \cite{Fiaschi14} and corresponds to a high-resolution (1400x1400) video with 9000 frames containing a large number of Drosophila larvae. As there is a limited movement between the upper and the lower halves, we used the upper half of the video for training and validation, and test on the lower half.

The \textbf{Coli} dataset contains 9 large (1024x1024) phase-contrast microscopy images of colonies of E.Coli bacteria (grown in mono-layer). Each image contains on average 514 worms. We use 7 images for training, 1 images to validate the method parameters, and report results on the remaining 2 images containing 531 worms. Unlike the first two datasets, the background segmentation task is not trivial, and we learn a linear pixel classifier using the convolutional features at the first layer of the SON-network for foreground/background segmentation.

\subsection{Rendering}
Our approach requires rendering a large number of crowded patches to train SON-networks and another collection of patches to form the gallery. Since the three datasets were of different kind and had different types of annotation, we used three different approaches to tackle these rendering tasks. For the C.Elegans case, we followed \cite{Wahlby10} and took the skeletons of singleton worms to build the PCA-based shape model that was further used to generate both crowded patches for SON-training as well as gallery patches. For the Larvae dataset, we simply isolated the singleton worms from the top of the dataset and used various similarity transforms and superimpositions to do the rendering (examples of renderings can be seen in \fig{teaser}). Finally, for E.Coli we do not have sufficient number of singleton bacteria that would be easy to isolate. However, the bacteria have simpler shapes mostly defined by the two endpoints (and were annotated this way). We therefore align worms of the same length rotated to a certain fixed orientation to define a certain class for SON-network training. We also created the gallery directly from training patches (for which the pose of the central worm is known). 

The sizes of the training datasets were 1000 classes with 500 images in each for C.Elegans and Larvae, and 198 classes of average size 550 for E.Coli. The gallery size was 4 millions for C.Elegans, 6 millions for Larvae and 240 thousands for E.Coli. Generally the patch sizes were 100x100, 40x40 and 50x50 for C.Elegans, Larvae and E.Coli respectively.

\subsection{Network design and training}

\begin{figure}
    \centering
    \includegraphics[width=0.46\textwidth]{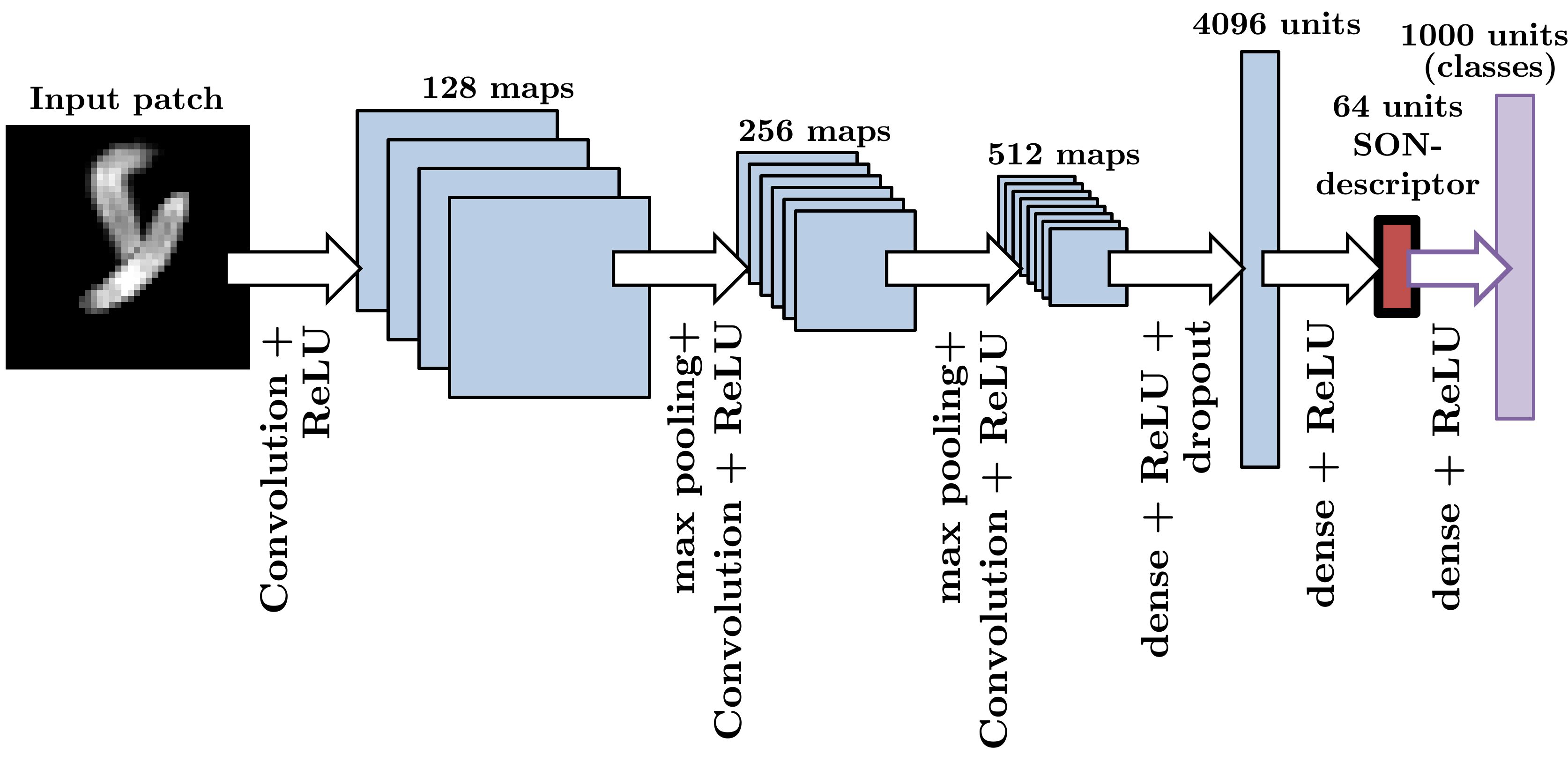}
    \caption{The architecture of the SON-networks used in our experiments. SON-descriptors are obtained by passing the input patch through two convolutional layers and two fully-connected layers (interleaved with rectified linear units (ReLU) and $2\times2$ max-pooling with downsampling layers. The network is trained so that the SON-descriptors could be used to linearly classify synthetic classes. See text for more details.}
    \label{fig:arch}
\end{figure}
In the experiments the SON-networks had the architecture specified in \fig{arch}. The SON-descriptors are thus 64 dimensional. For the Coli dataset the number of convolutional maps in each layer was halved.

To learn the classifiers the network was augmented at training time by an additional ReLU layer followed by a linear dense layers with 1000 (C.Elegans, Larvae) or 198 (E.Coli) output units corresponding to different classes. The classification was trained with the softmax loss. We used stochastic gradient descent algorithm (with momentum) and trained networks for 53, 40 and 100 epochs for Larvae, C.Elegans and E.Coli respectively. After each epoch the learning rate was scaled by factor 0.85.

\subsection{Matching with SON-descriptors}

\newlength{\lennine}
\setlength{\lennine}{4.2cm}

\begin{figure*}
    \centering
    \begin{tabular}{ccc}
    \includegraphics[width=\lennine]{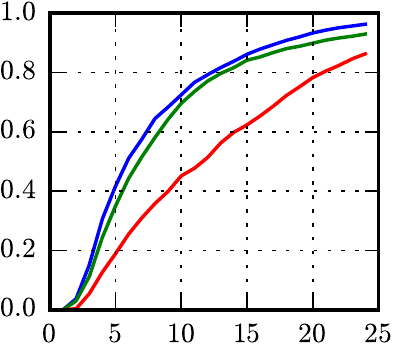}&\quad
    \includegraphics[width=\lennine]{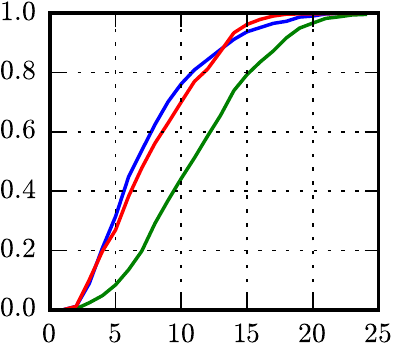}& \quad
    \includegraphics[width=\lennine]{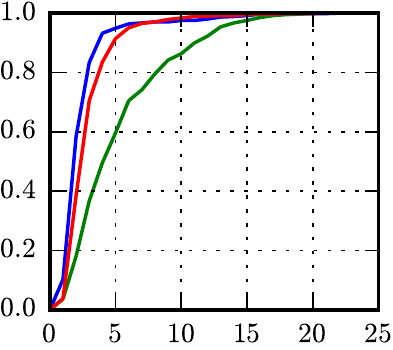}\\
    C. Elegans& Larvae & Coli
    \end{tabular}
    \caption{Cumulative plots of the accuracy of pose matching between randomly drawn query patches containing crowded organisms and a gallery of training patches. The accuracy is judged using symmetric Hausdorf distance between centerlines (using ground truth annotations). For matching, we compare distances between SON-descriptors ({\color{blue}blue line}), SIFT descriptors with optimally picked radii ({\color{red}red line}) and the $L_2$-distance between raw image patches ({\color{green}green line}). The plots show the number of samples (y-axis) with the pose distance less than threshold (x-axis). In all three datasets, the SON-distance yields better performance than SIFT and raw patches. }
    \label{fig:quant}
\end{figure*}

As our main contribution is the idea of hypothesis mining on the basis of SON-descriptors, we quantitatively compare the performance of the learned descriptors against SIFT descriptors \cite{lowe99} and a simple baseline based on L2 pixelwise distance. In more detail, on the test set we use $600-800$ patches crowded with organisms as queries, and then find their nearest neighbors in the gallery set. The query patches were sampled from the test set. Patches from C.Elegans and Larvae datasets were centered on a random pixel of random object from test set while E.Coli patches were chosen in a such a way that the center of each patch matched to the center of some segment which defines a E.Coli sample from test set. SIFT descriptors were calculated for the central pixel of each patch and with fixed keypoint orientation. We gave SIFTs an advantage by optimizing the diameters of keypoints neighborhoods on the test sets (separately for each dataset).

%\newlength{\lennine}
\setlength{\lennine}{1.3cm}

\newcommand{\inceleg}[1]{
\includegraphics[width=\lennine]{figures/patch_celegans/#1_input.png}&
\includegraphics[width=\lennine]{figures/patch_celegans/#1_output_L2.png}&
\includegraphics[width=\lennine]{figures/patch_celegans/#1_output_sift.png}&
\includegraphics[width=\lennine]{figures/patch_celegans/#1_output_nn.png}\\
}

\newcommand{\incecoli}[1]{
\includegraphics[width=\lennine]{figures/patch_ecoli/#1_input.png}&
\includegraphics[width=\lennine]{figures/patch_ecoli/#1_output_L2.png}&
\includegraphics[width=\lennine]{figures/patch_ecoli/#1_output_sift.png}&
\includegraphics[width=\lennine]{figures/patch_ecoli/#1_output_nn.png}\\
}

\newcommand{\inlarvae}[1]{
\includegraphics[width=\lennine]{figures/patch_larvae/#1_input.png}&
\includegraphics[width=\lennine]{figures/patch_larvae/#1_output_L2.png}&
\includegraphics[width=\lennine]{figures/patch_larvae/#1_output_sift.png}&
\includegraphics[width=\lennine]{figures/patch_larvae/#1_output_nn.png}\\
}

\begin{figure}
\centering
%\begin{tabular}{cccccccc}
\begin{tabular}{cccc}
\inceleg{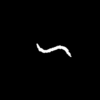}
\inceleg{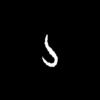}
\inceleg{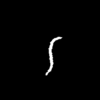}
\inceleg{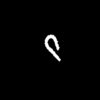}
%\inceleg{250}
%\inceleg{300}
query&$L_2$(raw)&SIFT&SON\\
\end{tabular}
\caption{Nearest neighbors in the gallery for the query patches using L2 distance on raw pixels, SIFT and SON-descriptors for the C.Elegans dataset.  \textit{Uniform sampling of test sets is shown.}}
\label{fig:patchworms}
\end{figure}

\begin{figure}
\centering
\begin{tabular}{cccc}
\incecoli{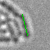}
\incecoli{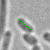}
\incecoli{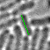}
\incecoli{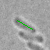}
%\incecoli{250}
%\incecoli{300}
%\incecoli{350}
%\incecoli{400}
query&$L_2$(raw)&SIFT&SON\\
\end{tabular}
\caption{Nearest neighbors in the gallery for the query patches using L2 distance on raw pixels, SIFT and SON-descriptors for the E.Coli dataset. Green lines are superimposed for clarity. \textit{Uniform sampling of test sets is shown.}}
\label{fig:patchcoli}
\end{figure}

\begin{figure}
\centering
\begin{tabular}{cccc}
\inlarvae{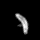}
\inlarvae{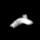}
\inlarvae{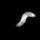}
\inlarvae{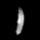}
%\inlarvae{250}
%\inlarvae{300}
%\inlarvae{450}
%\inlarvae{400}
query&$L_2$(raw)&SIFT&SON\\
\end{tabular}
\caption{Nearest neighbors in the gallery for the query patches using L2 distance on raw pixels, SIFT and SON-descriptors for the Larvae dataset.  \textit{Uniform sampling of test sets is shown.}}
\label{fig:patchlarvae}
\end{figure}

When searching for the closest neighbors we either use L2-pixelwise distance or distances based on SIFT or  SON-descriptors. In all cases, we evaluate the distances between the poses of the central organism in the query patch and the closest nearest neighbor. To measure the disparity between the poses we used the symmetric Hausdorf distance ($d_H(A,B) = \max\{  \max_{a\in A} \min_{b \in B} \|a-b\|, \max_{b\in B} \min_{a \in A} \|a-b\| \}$) between centerlines.

\fig{quant} provides numerical comparisons, while \fig{patchworms}, \fig{patchcoli}, \fig{patchlarvae} provides side-by-side qualitative comparisons. For all three datasets, the SON-descriptors provide more accurate pose matches compared to SIFT and L2-distance between patches. When comparing our method with L2-distance-based method, it is important to note that matching using SON-descriptors is much faster as its dimensionality (set to 64 in all our experiments) is much smaller than the dimensionality of image patches. For SIFT descriptors the best results were achieved on keypoints neighborhood of size 5 pixels for E.Coli and C.Elegans and 4 pixels for Larvae. Because of low variability of Larvae and E.Coli such a small neighborhood can provide a good estimation for the whole organism.

\subsection{Image parsing}

Finally, we provide results for full image parsing on our dataset. Numerically, we compare the results on C.Elegans and Larvae with the results of \cite{Wahlby10} (using their \texttt{WormToolbox} software). Their method is specialized for C.Elegans and obtains near-perfect results on their dataset. 
%The method exploits approximate constancy of worm sizes, and for each cluster pre-estimates the size of any cluster that they find prior to parsing it. We therefore consider a variant of our method that performs the same pre-estimation, and uses the estimate in the following way. If the cluster is a singleton, it naturally declares the whole segment a single worm without further processing (a simple skeletonization can then estimate the pose). If the cluster has more than one worm, we use the estimated number as a constraint on the number of active hypotheses within our parsing model (the corresponding equality is simply added to the integer program). 

\begin{figure}
    \centering
    \includegraphics[height=4.5cm]{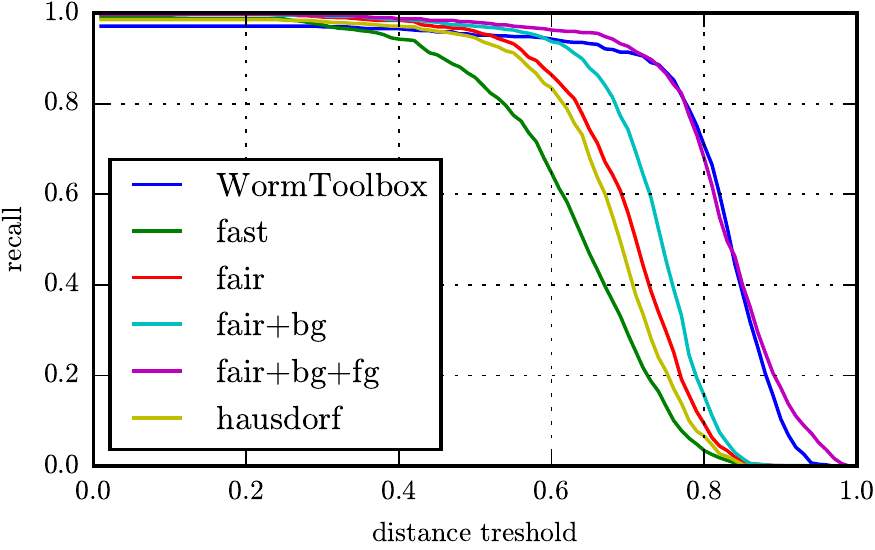}\\
    %\caption{The graph shows the ratio of true positives (y-axis) lying within the certain Hausdorf distance threshold (x-axis) from the organisms found by the methods (all three methods produced similar number of organisms). At most one-to-one matching is enforced using Hungarian algorithm. For C.Elegans, the performance of the proposed approach is mostly below the performance of WormToolbox \cite{Wahlby10}, which is specialized for this data due to lower accuracy of localization (lower values of threshold). Good performance of the proposed method for large thresholds suggests that it can benefit from local optimization of the poses.}
    \caption{The graph shows the ratio of true positives (y-axis) lying within the certain intersection-over-union (IoU) score threshold (x-axis) with the organisms found by the methods (all methods produced similar number of organisms). At most one-to-one matching is enforced using Hungarian algorithm. For C.Elegans, the performance of the proposed approach without postprocessing ({\tt fast}, {\tt fair} and {\tt hausdorf} lines) is mostly below the performance of WormToolbox \cite{Wahlby10} due to lower accuracy of localization (large values of threshold). But with simple postprocessing steps the approach performs slightly better then WormToolbox.}
    \label{fig:celegans_plot}
\end{figure}

%In \fig{celegans_plot}, we compare the performance of the Worm toolbox of \cite{Wahlby10} with several variants of our method on C.Elegans. The variants differ in the way of definition $d_{ij}$ of optimization problem \eq{fl}. We refer as {\tt fast} to the variant with equation \eq{dij2} and {\tt fair} to the variant with \eq{dij1}. The {\tt hausdorf} variant is the variant where we estimate $d_{ij}$ via the Hausdorf distance between centerlines of the worm hypotheses suggested for the $i$th and the $j$th pixel. Each variant is also computed with and without cluster size pre-estimation. Generally, our method here performs worse than WormToolbox for some thresholds and slightly better for others. 
In \fig{celegans_plot}, we compare the performance of the Worm toolbox of \cite{Wahlby10} with several variants of our method on C.Elegans. The variants differ in the way of definition $d_{ij}$ of optimization problem \eq{fl}. We refer as {\tt fast} to the variant with equation \eq{dij2} and {\tt fair} to the variant with \eq{dij1}. The {\tt hausdorf} variant is the variant where we estimate $d_{ij}$ via the Hausdorf distance between centerlines of the worm hypotheses suggested for the $i$th and the $j$th pixel. Generally, all three variants performs worse than WormToolbox, especially for large thresholds. However these results can be significantly improved by local fitting of the pose. In order to show it we performed two simple postprocessing steps on {\tt fair} results. On the first step we excluded from instances masks all pixels which belong to background of initial image (we refer to this results as {\tt fair+bg}). On the second step we assigned all uncovered foreground pixels of the image to the nearest instances masks ({\tt fair+bg+fg} line in \fig{celegans_plot}). With this postprocessing our approach slightly outperforms WormToolbox on the most of thresholds of intersection-over-union (IoU) score.

The main limitation of method proposed in \cite{Wahlby10} is the ability to work only with binary masks of worms. Therefore we trained WormToolbox model for Larvae using the binary masks of the singleton worms (which were also used in our method for training SON-network) and tested their model on binary masks of worm clusters. Since our method is able to work with any type of input images, we trained and tested it on actual grayscale images. We also considered exploiting our prior knowledge about data: Larvae organisms usually tend to overlap significantly. That fact makes choosing the good set of hypotheses ambiguous. To facilitate picking smaller hypotheses, we change $\lambda$ in \eq{fl} to $\lambda_0 + \lambda_1 s_i$, where $s_i$ is the size of hypothesis $h_i$. Table~\ref{tab:larvae_results} provides precision, recall and F1-score values for two methods on Larvae. The big advantage of our method over the Worm toolbox is likely due to the use of textural information that is important for parsing organisms clusters whose overlapping area size is comparable with the size of the whole organism. 

We also present qualitative results in \fig{qualitative}. Both qualitative and quantitative results demonstrate the ability of our method to parse complex clusters. The method gets limited accuracy in terms of finding the exact worm boundaries, which can be addressed through local shape/pose optimization using our result as the initial starting point.

\begin{table}
    \centering
    \begin{tabular}{r | c |c|c}
         & Precision & Recall & F1-score \\ \hline
         WormToolbox & 0.7032 & 0.4665 & 0.5609 \\[0.1cm] %\hline
         SON fast & 0.8174 & 0.8302 & 0.8238 \\
        SON fast + size & 0.8291 & 0.8787 & 0.8532 \\[0.1cm] %\hline
        SON fair & 0.8091 & 0.8103 & 0.8097 \\
        SON fair + size & 0.8340 & 0.8317 & 0.8329 \\[0.1cm] %\hline
        SON hausdorf & 0.8277 & 0.8431 & 0.8353 \\ 
        SON hausdorf + size & \textbf{0.8816} & \textbf{0.8816} & \textbf{0.8816} \\ \hline
    \end{tabular}
    \caption{Precision, recall and F1-score on Larvae dataset. Methods results were evaluated by Hausdorf-distance-based Hungarian matching with threshold set to mean thickness of larvae (8 pixels).}
    \label{tab:larvae_results}
\end{table}

\newlength{\third}
\setlength{\third}{4.6cm}

\begin{figure*}[t]
    \centering
    \begin{tabular}{ccc}
    \includegraphics[width=\third]{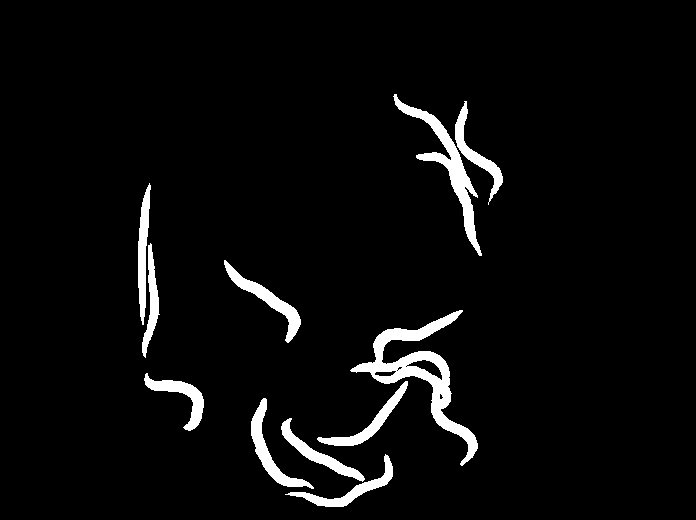}&
    \includegraphics[width=\third]{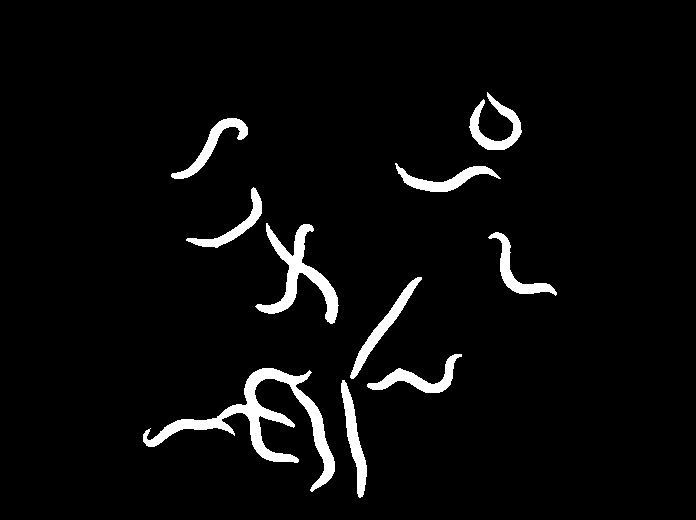}&
    \includegraphics[width=\third]{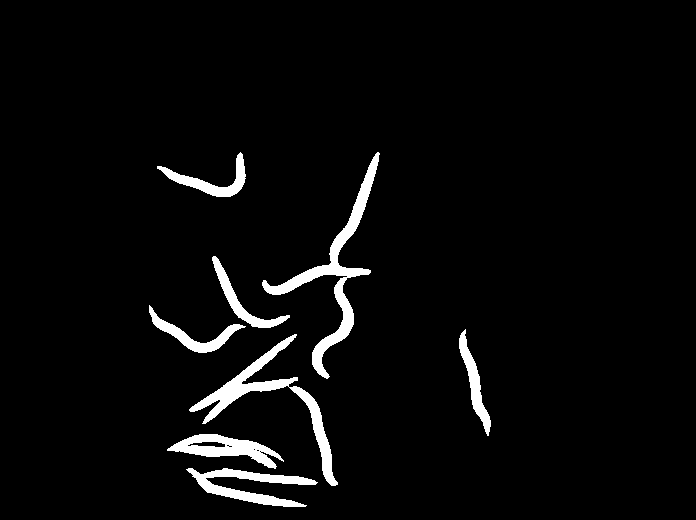}\\
    \includegraphics[width=\third]{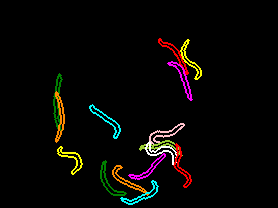}&
    \includegraphics[width=\third]{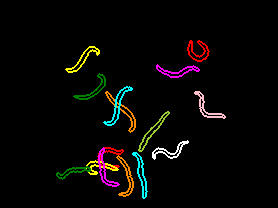}&
    \includegraphics[width=\third]{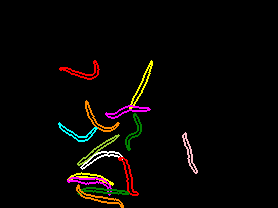}
    \end{tabular}\\
    (a) Parsing (bottom row) of segmented wells (top row) with C.Elegans 
    
    \vspace{1mm}
    
    \includegraphics[width=\textwidth]{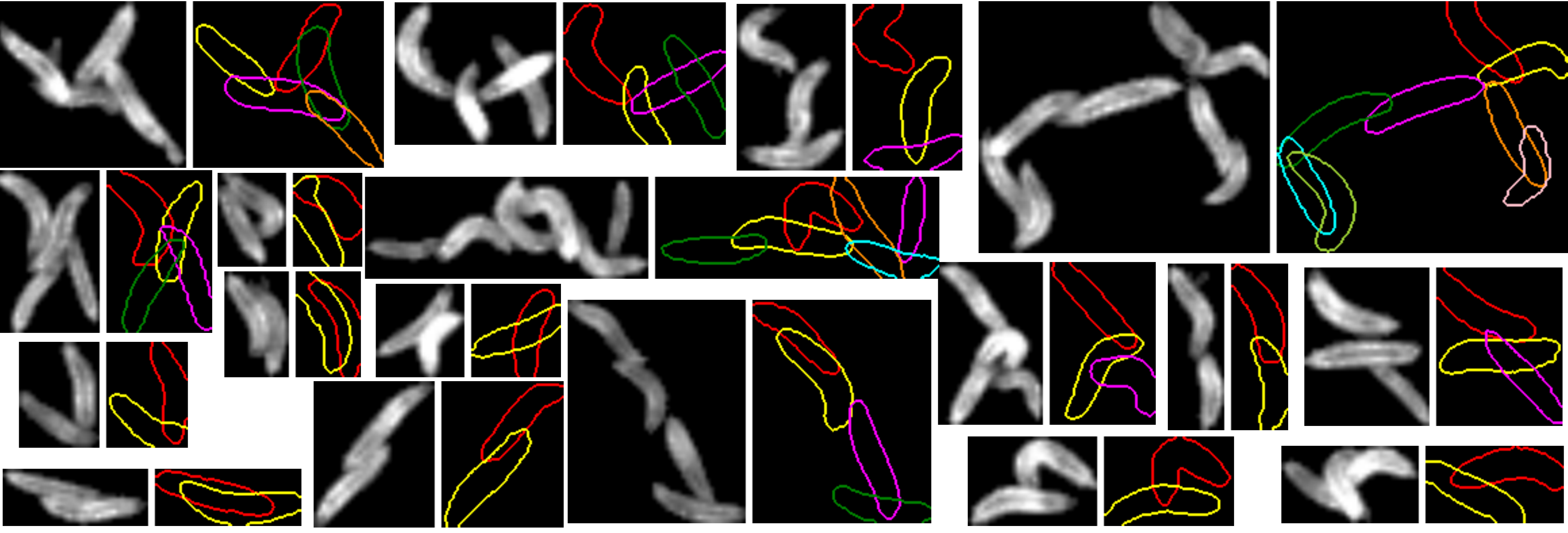}
    (b) Parsing of clusters of Drosophila larvae (left -- input cutout, right -- results)
   
  %    \includegraphics[width=\textwidth]{figures/ecoli_results.pdf}
 %   (c) Parsing of E.Coli clusters (left -- input cutout, right -- results)
    
    \caption{Qualitative results (randomly chosen) for the C.Elegans and the Larvae datasets. Better viewed in color and with zoom-in. We used {\tt fair} variant without postprocessing for C.Elegans and {\tt hausdorf} variant with modification in \eq{fl} for Larvae.}
    \label{fig:qualitative}
\end{figure*}

%% file: conclusion.tex
\section{Summary}
\label{sect:discussion}

We have presented a new approach to parsing images with crowded objects and have demonstrated its viability for biomedical images of model organisms with medium complexity. Our approach is not specific to these kinds of objects and can be applied to other data. The method assumes that it is possible to render training data (although in the case of E.Coli we showed that one can obtain training data directly from user-annotated images). Another assumption is that the gallery can sample the pose space of a single object densely enough.  In the current version, we also assume that foreground/background segmentation can be performed as a pre-processing.

Our main contribution is in the mechanism for proposing hypotheses about individual objects that is based on singling-out networks. This mechanism is compatible with different hypotheses selection approaches. E.g.\ it can be used within full-fledged ``analysis-by-synthesis'' approach that would choose a subset of hypotheses that minimizes the mismatch in appearance between the synthetic and the real image directly. In this work, we presented a faster discrete optimization-based alternative that produced good results in our experiments.